\documentclass[a4,notitlepage,12pt]{jedm}

\usepackage{times}
\usepackage{amsfonts}
\usepackage{graphicx} 
\usepackage{subfigure} 
\usepackage{booktabs}

\usepackage{algorithm}
\usepackage{algorithmic}
\usepackage[cmex10]{amsmath}
\usepackage{xcolor}
\usepackage{placeins}

\renewcommand{\thesection}{\arabic{section}}
\renewcommand{\thesubsection}{\thesection.\arabic{subsection}}
\renewcommand{\thesubsubsection}{\thesection.\arabic{subsection}.\arabic{subsubsection}}

\newcommand{\NORD}{SPRITE}

\begin{document}
\title{SPRITE: A Response Model \\ For Multiple Choice Testing}


\author{{\large RYAN NING } \\Rice University\\ryan.ning@sparfa.com \and {\large ANDREW E. WATERS}\\OpenStax \\waters@sparfa.com\\[0.5cm] \and {\large CHRISTOPH STUDER} \hspace{3.55cm} \\ Cornell University \\studer@sparfa.com \and {\large RICHARD G. BARANIUK}\\Rice University \\richb@sparfa.com}
\date{}

\maketitle
\begin{abstract}

\noindent Item response theory (IRT) models for categorical response data are widely used in the analysis of educational data,  computerized adaptive testing, and psychological surveys. However, most IRT models rely on both the assumption that categories are strictly ordered and the assumption that this ordering is known a priori. These assumptions are impractical in many real-world scenarios, such as multiple-choice exams where the levels of incorrectness for the distractor categories are often unknown.  While a number of results exist on IRT models for unordered categorical data, they tend to have restrictive modeling assumptions that lead to poor data fitting performance in practice. Furthermore, existing unordered categorical models have parameters that are difficult to interpret. In this work, we propose a novel methodology for unordered categorical IRT that we call SPRITE (short for stochastic polytomous response item model) 
that: (i) analyzes both ordered and unordered categories, (ii) offers interpretable outputs, and  (iii) provides improved data fitting compared to existing models. 
%
We compare SPRITE to existing item response models and demonstrate its efficacy on both synthetic and real-world educational datasets.

\end{abstract}






\section{Introduction}
\label{sec:introduction}

{

 }  
{\noindent
\subsection{Item Response Theory}
A common task in education is evaluating how well learners in a class have mastered some set of competencies.  This task is almost universally carried out through some form of testing, typically where a student is given a set of questions, and their ability is measured simply by counting the number of questions they answer correctly. This simple method of counting the number of correct responses is called classical test theory (CTT) \cite{bechger2003using}. However, CTT ignores valuable information in the way respondents interact with each question. For example, two respondents can have the same number of correct responses but have completely different areas of mastery---information that cannot be modeled by the simple aggregate score.
\par Item response theory (IRT), in contrast, models the interaction between each item\footnote{The term ``item,'' in IRT, is a general term for any response item. In the education and testing domain, the term "item" corresponds to a test or homework question.} and respondent to learn information regarding respondent and item characteristics \cite{nering2011handbook}. Concretely, IRT explicitly models the probability that a respondent will choose each multiple choice category (option) within a question. IRT is widely considered to be superior than classical test theory in both efficiency and accuracy to achieve high precision in measuring a respondent's characteristics with a smaller number of questions \cite{nering2011handbook}. In recent years, IRT has seen wide-spread adoption in analyzing surveys, questionnaires, and standardized tests, such as the Graduate Record Examination (GRE) and Graduate Management Admission Test (GMAT) \cite{ware2000practical}.  

}

\subsection{The Problem of Unordered Categories}
Data typically analyzed by IRT may have ordered or unordered categories.\footnote{The term ``categories,'' commonly used in the IRT community, refers to the multiple choice category labels within each question. These categories may be ordered in a meaningful way and should not be confused with strictly categorical data, where no particular ordering of the categories exists \mbox{\cite{agresti2002categorical}}.} These categories may be ordered on a scale (such as a survey questionnaire, where respondents are asked to provide an answer on a scale from one to five) or they can be ordered in more abstract ways, such as the correctness of a response to a test question. Most IRT models rely on two key assumptions: 1) the categories are strictly ordered and 2) this ordering is known a priori. These assumptions are impractical in many real-world scenarios. For example, in a multiple-choice testing question with no strictly ordered categories, there may be a correct category and multiple distractor categories that are equally incorrect.  Furthermore, even with ordered categories, the category ordering itself may not be known in advance. As a concrete example, assume the following multiple-choice question: What is the capital of Brazil? A) S$\tilde{\text{a}}$o Paulo, B) Belo Horizonte, C) Beijing, or D) Brasilia. Category D) is the correct answer; categories A) and B) are not correct but these cities are both in Brazil and hence, these two categories can be considered to be equally wrong; category C) is the worst choice since Beijing is not in South America.  In this case, since categories B) and C) are equally incorrect, a strict ordering of the categories does not exist. Furthermore, even the non-strict ordering of the categories is often not known a priori unless a domain expert is providing this information (a costly procedure). To model such unordered\footnote{We use the term "unordered" to refer to categories with no ordering, including categories with partial ordering and ordered categories with a priori unknown category ordering.} categories, we need IRT models designed for unordered data.

 \newcommand{\nwidth}{0.01}
 \newcommand{\nheight}{0.01}
 
 \renewcommand{\NORD}{SPRITE}
 \begin{figure*}[t]
\begin{center}
   \subfigure[A set of three Gaussian SPRITE functions or ``sprites.'']{\label{sprite_func}\includegraphics[width=0.2\columnwidth]{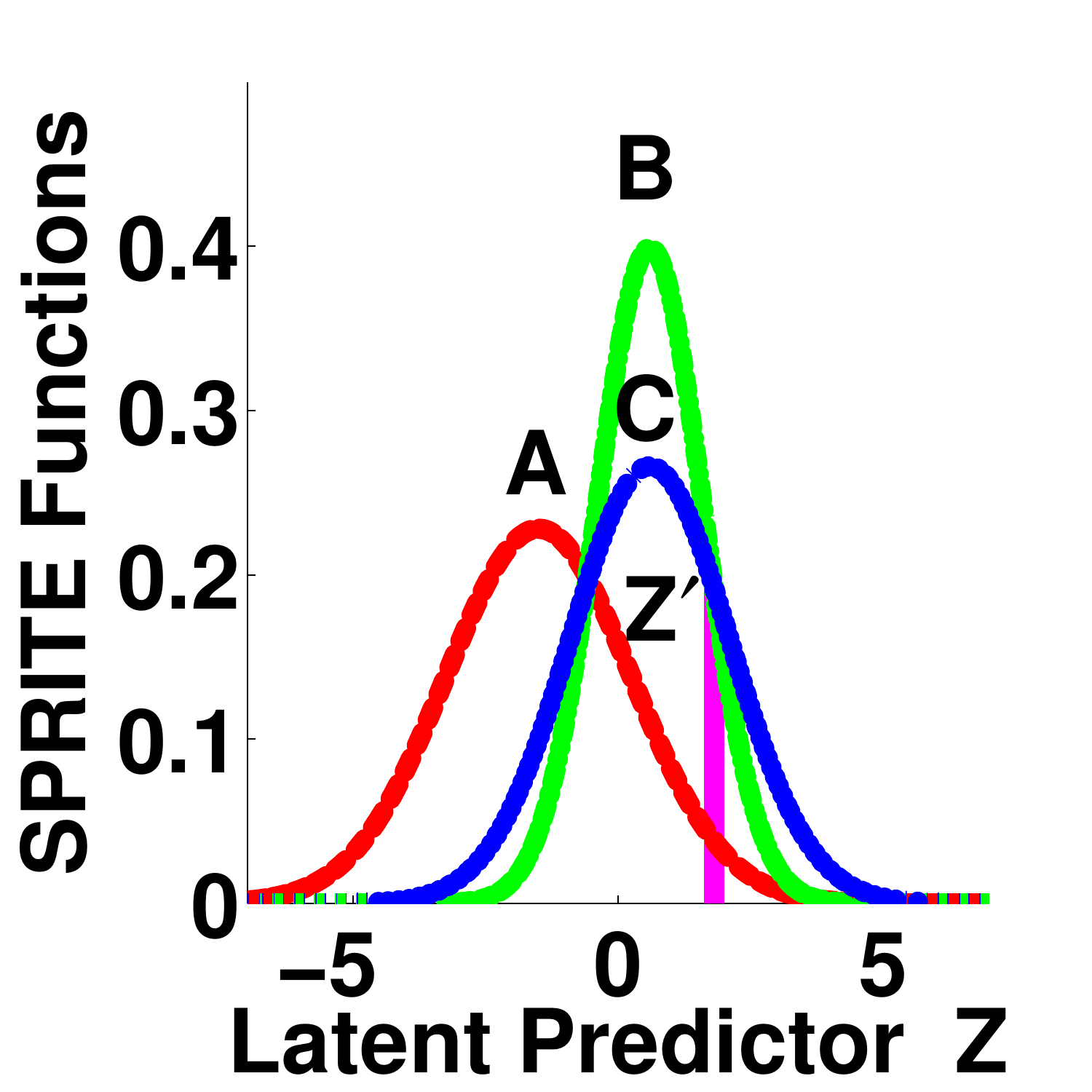}}\hfill
    \subfigure[Probabilistic model underlying SPRITE.]{\label{sprite_likelihood}\includegraphics[width=0.2\columnwidth]{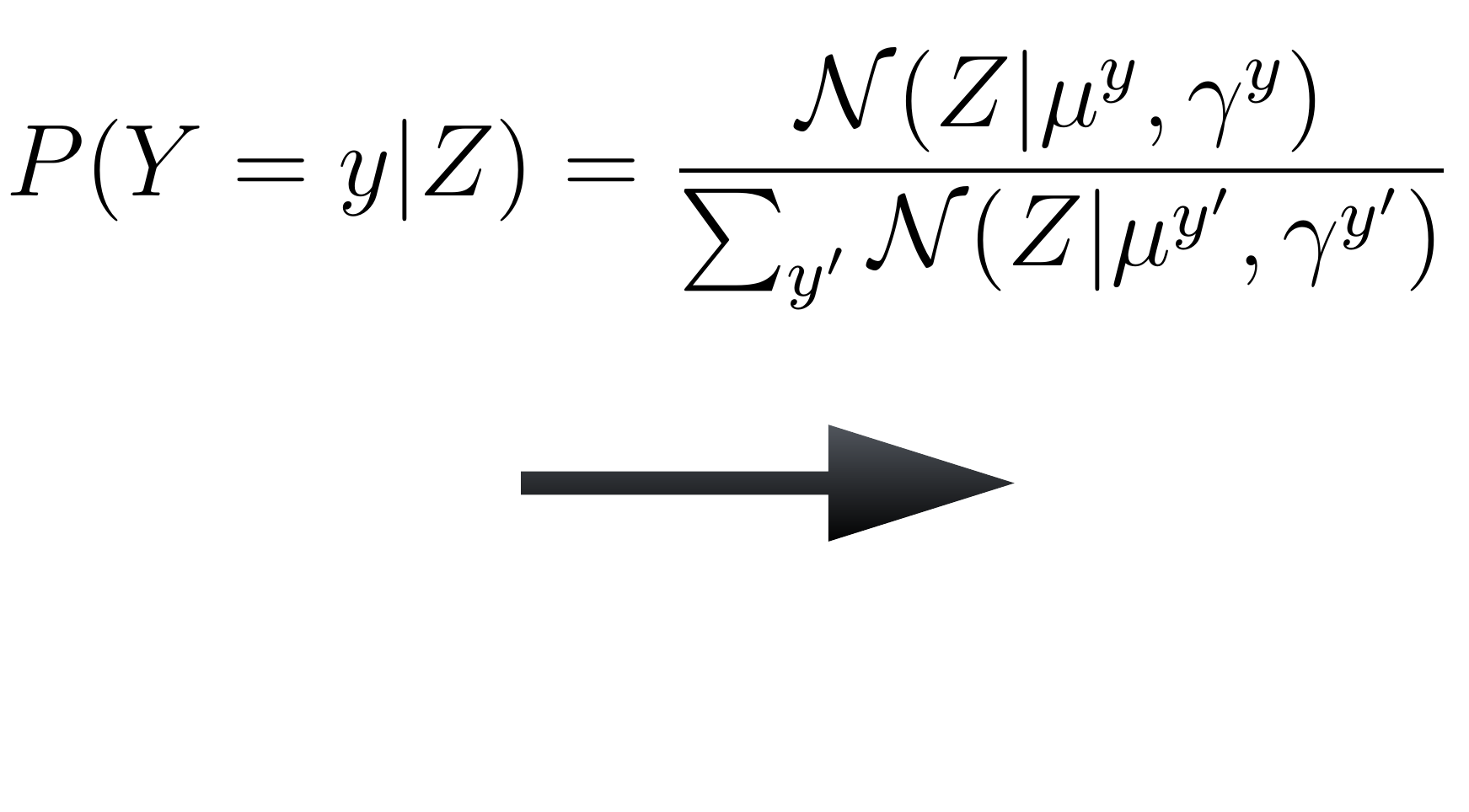}} \hfill
    \subfigure[SPRITE choice probabilities or item category response functions.]{\label{sprite_icrf}\includegraphics[width=0.2\columnwidth]{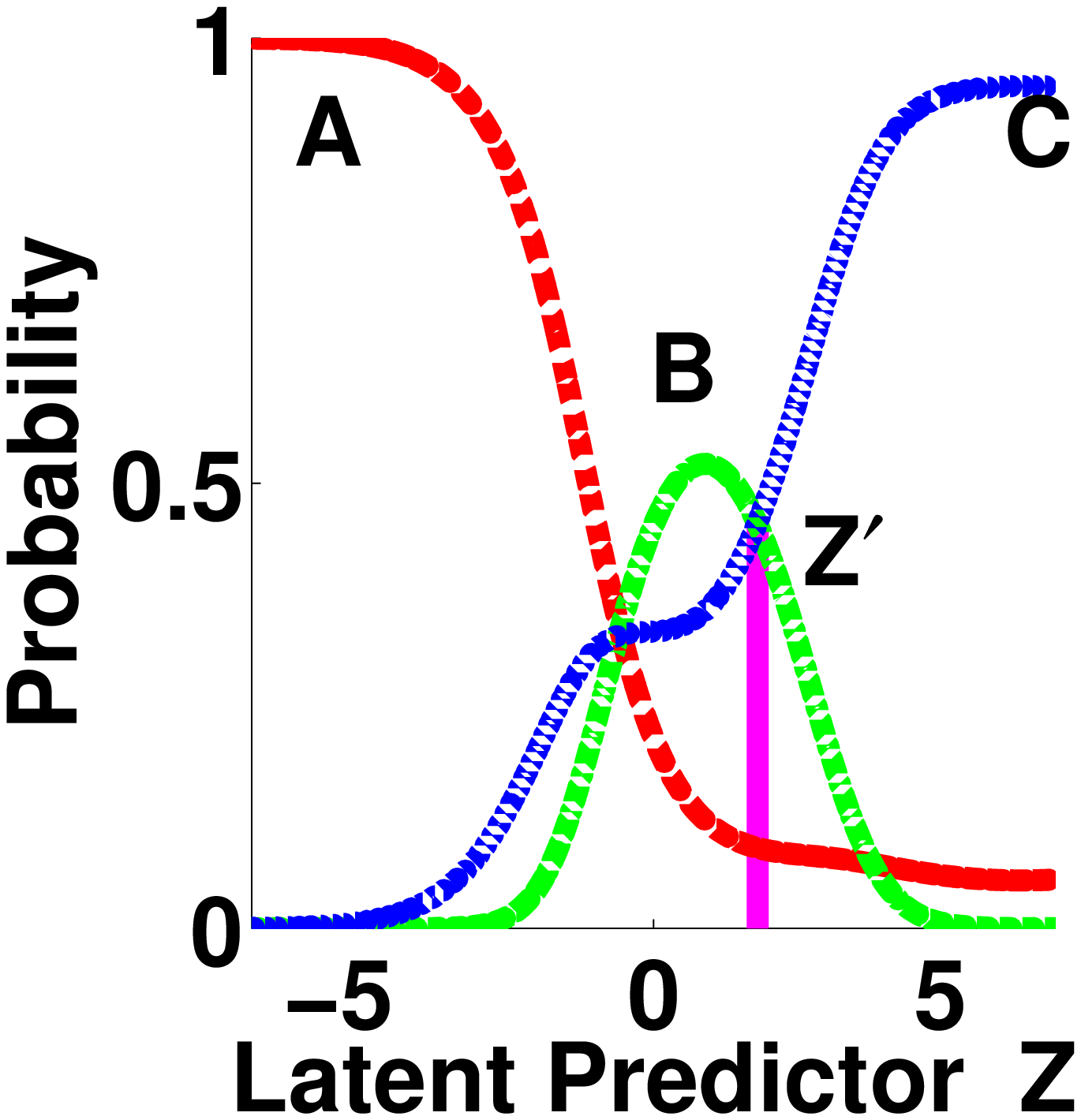}} \hfill
    \subfigure[Category probabilities for latent trait $Z^\prime$.]{\label{sprite_bar}\includegraphics[width=0.2\columnwidth]{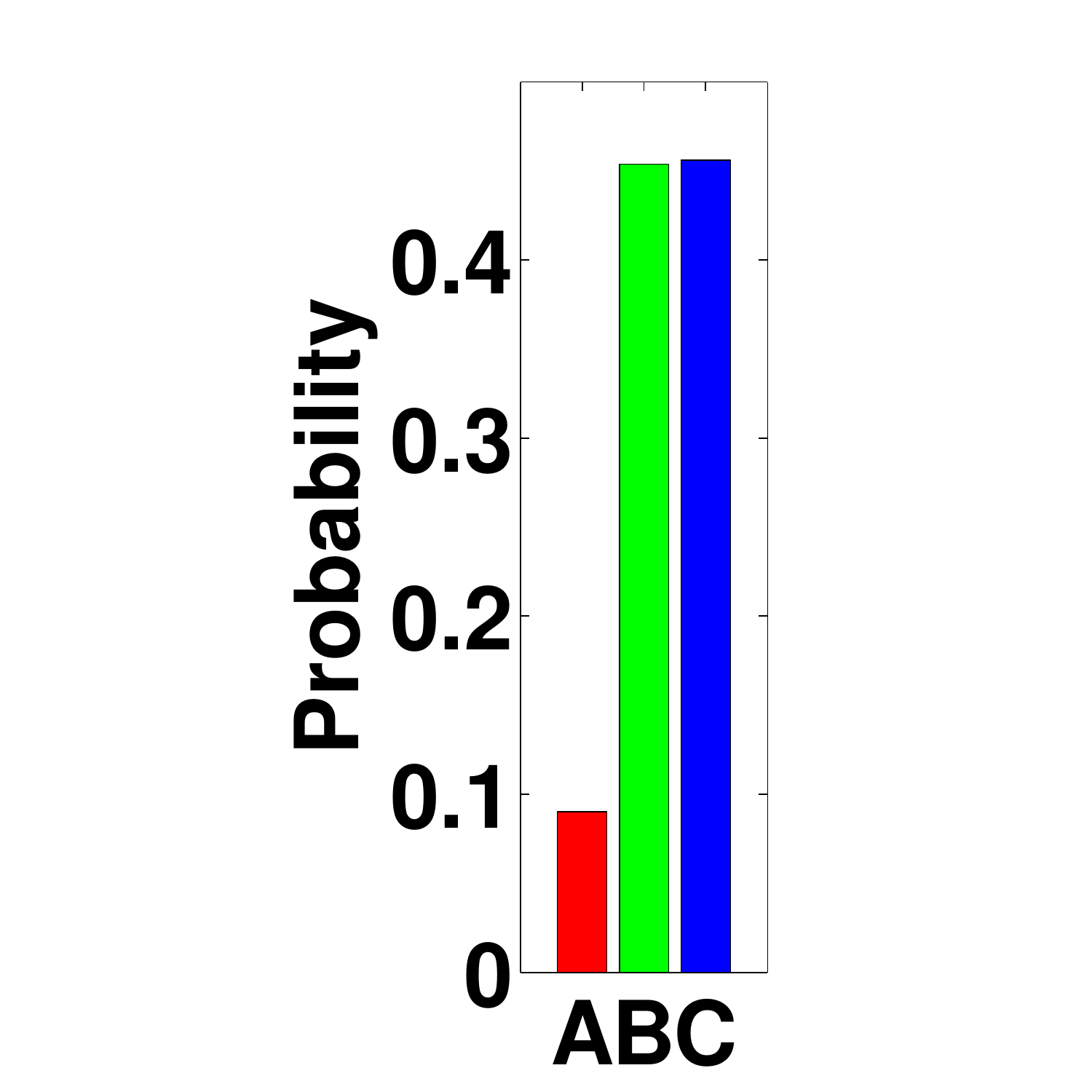}}
\end{center}
\caption{Illustration of the SPRITE model: The location of the latent-predictor variable $Z$ and the Gaussian category functions (referred to as ``sprites'') induce a probability mass function determining the likelihood of the category $y$ (out of A, B, and C) a respondent will choose. (a)~The Gaussian functions or sprites associated with each category. (b) The SPRITE likelihood model. (c) The resulting choice probabilities as a function of $Z$. These curves are called item category response functions (ICRFs) by the IRT community. (d) The category choice probabilities for a particular latent-predictor value of $Z^\prime$ indicated by the vertical line in (a) and (c). The flexibility of SPRITE enables us to model ordered as well as unordered categories.}
 \label{fig:sprite_overview}
\end{figure*}

 \subsection{Prior Art}

Some IRT models for unordered data, such as the generalized partial credit model (GPCM) \cite{muraki1992generalized}, have restrictive modeling assumptions that degrade data-fitting performance in practice. Other IRT models for unordered data, such as the nominal response model (NRM) \cite{bock1972estimating}, produce output parameters that are hard to interpret\cite{thissen1997response}. In many applications, interpretability of the posterior model parameters provide insight into the underlying characteristics of the data. Thus, having interpretable output parameters is often critical. We discuss existing IRT models for unordered data in detail in Section \ref{sec:irt}.

An alternative class of models that have not been widely adopted in the IRT community are Bayesian ordinal models. State-of-the-art Bayesian methods \cite{odm99} for modeling ordinal data generally rely on the assumption of a discrete set of ordered bins which, when combined with a latent predictor variable, induce a probability distribution over the set of ordered categories. We refer to this particular model as the ORD (short for ORDinal) model in the following. However, the ORD model can only be deployed on datasets with a known, strict ordering. To enable inference on unordered datasets, we make a slight modification to the ORD model to allow for unordered response data and refer to it as LORD (short for learned ordinal). We discuss ORD and LORD in detail in Section \ref{sec:ORD}.

%
%
%
\subsection{Contributions}

We propose a novel IRT model for unordered categorical data that we dub SPRITE (short for stochastic polytomous response item model). 
For each respondent, SPRITE directly models the probability of choosing each category over the respondent's latent parameter space.    
An illustration of SPRITE is shown in Figure~\ref{fig:sprite_overview}. SPRITE produces meaningful category orderings and enables the analysis of both ordered and unordered categorical response data. In addition, SPRITE offers a high level of interpretability and provides statistics on the informativeness of questions and categories. Lastly, SPRITE provides (often significantly) better data-fitting performance  than existing IRT models for unordered data. \par Table \ref{tbl_results} demonstrates the superiority of SPRITE for a set of real-world datasets in terms of prediction performance. The details about the individual data sets are summarized in Table \ref{tbl_datasets}. In our experiments, we compare the data fitting ability of SPRITE against existing unordered IRT models by looking at predictive performance. We withhold 20\% of the observed responses from the data and impute the missing responses using each model. We compute the prediction error rates as the number of false  predictions over the total number of predictions and see that SPRITE outperforms the competing models for all considered datasets. 

\begin{table}[t]
\centering
\caption{Prediction error (the number of false predictions over the total number of predictions) and standard deviation of SPRITE, ORD \protect\cite{odm99}, 
NRM \protect\cite{bock1972estimating}, and GPCM \protect\cite{muraki1992generalized} on various datasets. For ORD, the superscript $^\text{L}$ indicates that we used a modified version of the standard Bayesian ordinal model, where we learn the category orderings directly from data, which are unavailable for that particular dataset (see Section \ref{sec:ORD} for the details).  SPRITE obtains the best prediction performance on all datasets. \label{tbl_results}}{
\begin{tabular}{lcccc}
\toprule
%
\textbf{Description} & \textbf{SPRITE} & \textbf{(L)ORD}  & \textbf{NRM} & \textbf{GPCM} \\
\midrule
Algebra test & \textbf{0.25 (0.01)}& 0.29 (0.02)$\,\,\,$ &0.31 (0.01)&0.26 (0.01)\\
Computer engineering course & \textbf{0.17 (0.01)} & 0.31 (0.02)$^\textbf{L}$&0.33 (0.01)&0.21 (0.01)\\
Probability course & \textbf{0.41 (0.01) }& 0.68 (0.03)$^\textbf{L}$ &0.57 (0.01)&0.62 (0.01)\\
Signals and systems course & \textbf{0.29 (0.01)}& 0.48 (0.02)$^\textbf{L}$&0.41 (0.01)&0.56 (0.01)  \\
Comprehensive university exam & \textbf{0.53 (0.01)}& 0.71 (0.01)$^\textbf{L}$&0.63 (0.01)&0.62 (0.01)  \\
\bottomrule
\end{tabular}}
\end{table}

%
%
%
%
\subsection{Paper Outline}

This paper is organized as follows. In Section~\ref{sec:model}, we review IRT modeling and prior art for unordered categorical IRT. In Section~\ref{sec:sprite_model}, we introduce the SPRITE model. In Section~\ref{sec:method}, we develop a Markov Chain Monte--Carlo (MCMC) sampling method for SPRITE. In Section~\ref{sec:experiments}, we present results for both synthetic and real-world data experiments. We conclude in Section~\ref{sec:conclusion} with a summary and directions for future research.


\section{Existing Statistical Models For Item Response Theory}
\label{sec:model}
We describe our notation and the IRT modeling assumptions in Section \ref{sec:model_parameters}. We present existing IRT models in Section \ref{sec:irt} and existing Bayesian ordinal models in Section \ref{sec:ORD}.

\subsection{IRT Notation and Modeling Assumptions}
\label{sec:model_parameters}
Assume that we have a dataset consisting of $N$ respondents (for example, test takers on an exam) interacting with $Q$ questions (for example, multiple-choice questions in an educational scenario). The observed data matrix $\mathbf{Y}$ consists of all the observed interactions between respondents and questions with $Y_{ij}$ denoting the interaction result between the $i^\text{th}$ respondent and the $j^\text{th}$ question.  We assume that one observes polytomous (i.e., more than two categories per question) interaction data, i.e., $Y_{ij} \in \{1,\ldots,M_j\}$, where $M_j$ denotes the number of categories for question $j$. We allow the number of categories $M_j$ to vary across different questions. In many practical scenarios not every interaction ${Y}_{ij}$ is observed. Consequently, let $\Omega_{\text{obs}}$ define the index set of entries for which we observe data. \par We now wish to model the observed outcomes $Y_{ij} \in \Omega_\text{obs}$ in a statistically principled way.  We will assume a linear predictor $Z_{ij}$ that induces a discrete probability distribution over the set of $M_j$ categories. There are many models available for defining the predictor $Z_{ij}$, including linear regression \cite{gelman_book,bishop2006pattern}, low-rank models \cite{recht2010guaranteed,zhou2010nonparametric,lan2013sparse}, and cluster-based models \cite{busse2007cluster}. 
In this work, we will confine ourselves to Rasch-type models \cite{rasch1993probabilistic}. We focus on the Rasch model both for its simplicity, as well as for its applicability to a wide variety of ordinal data-modeling problems \cite{rasch1993probabilistic,pallant2007introduction,bond2013applying}. We note that our proposed models can easily be combined with more advanced predictor models such as multi-dimensional IRT (MIRT)  \cite{beguin2001mcmc} and sparse factor-analysis techniques \cite{lan2013sparse}.
Put simply, the Rasch model defines a random effect $\theta_i \in \mathbb{R}$ for all respondents $i=1,\ldots,N$, as well as a random effect $\alpha_j \in \mathbb{R}$ for all questions $j=1,\ldots,Q$. The linear predictor variable $Z_{ij}$ is then given by $Z_{ij} = \theta_i - \alpha_j$. In an educational context, $\theta_i$ corresponds to $i^\text{th}$ learner's ability and $\alpha_j$ corresponds to $j^{th}$ question's intrinsic difficulty.   
%

\subsection{Existing IRT Models}
\label{sec:irt}

IRT can be viewed as a generalized latent variable modeling technique. 
%
While numerous models for IRT exist in the literature, only a few of them are suitable for unordered categorical response data. In particular, GPCM \cite{muraki1992generalized} and NRM \cite{bock1972estimating} can be used to analyze unordered categorical response data. We now describe each existing IRT model in detail.

\subsubsection{Generalized Partial Credit Model (GPCM)}
\label{sec:gpcm}

The GPCM is a generalized version of the strictly ordinal partial credit model \cite{masters1982rasch} that allows partial ordering of the categories \cite{muraki1992generalized}. GPCM is constructed from successive dichotomization of adjacent categories. Under GPCM, the probability that respondent $i$ will choose category $y$ for question $j$ is given by
\begin{align}
P(Y_{ij} = y | \theta_{i}, \beta_{j}, \boldsymbol{\alpha_{j}}) = \frac{\text{exp}\big(\sum_{v=1}^{y}\beta_{j}(\theta_{i}-\alpha_{jv})\big)}{\sum_{k=1}^{M_j}\text{exp}\big(\sum_{v=1}^{k}\beta_{j}(\theta_{i}-\alpha_{jv})\big)},\label{eq:gpcm_likelihood}
\end{align}
\noindent where $\beta_{j} $ is a single discrimination factor for the $j^\text{th}$ question and $\boldsymbol{\alpha_{j}} = [\alpha_{j1}; \ldots; \alpha_{jM_j}]$ represents threshold values where adjacent categories have equal probability of being chosen. We refer the reader to \cite{muraki1992generalized} for the derivation of \eqref{eq:gpcm_likelihood}. Although GPCM can be used to analyze unordered categorical responses, the model still tries to impose a strict ordering of the categories.  Intuitively, GPCM says that the probability of choosing category $y$ is proportional to the probability of successively choosing $y$ adjacent pairs of categories (i.e., for a respondent to choose category~3, they have to first choose category 2 over category 1, and then choose category 3 over category 2, and finally choose category 3 over category 4). This construction of successive binary choices does not allow overlapping of the categories. As a result of this restrictive modeling assumption, GPCM does not perform well under conditions where the categories overlap.

\subsubsection{Nominal Response Model (NRM)}
\label{sec:nrm}

The NRM \cite{bock1972estimating} is suitable for modeling categorical response data with no particular order. NRM is useful when multiple categories are equally good or the ordering of categories is unknown. NRM uses independent exponential functions to model each categorical response. For NRM, the probability that respondent $i$ will choose category $y$ for question $j$ is given by
\begin{align*}
P(Y_{ij} = y | \theta_{i}, \boldsymbol{\beta_{j}}, \boldsymbol{\alpha_j}) = \frac{\text{exp}\big(\beta_{jy}(\theta_{i}-\alpha_{jy})\big)}{\sum_{k=1}^{M_j} \text{exp}\big(\beta_{jk}(\theta_{i}-\alpha_{jk})\big)}, 
\end{align*}
\noindent where $\boldsymbol{\beta_{j}} = [\beta_{j1}; \ldots; \beta_{jM_j}]$ is a vector of discrimination factors for the categories in the $j^\text{th}$ question and $\boldsymbol{\alpha_{j}} = [\alpha_{j1}; \ldots; \alpha_{jM_j}]$ is a vector of intrinsic difficulties for the categories in the $j^\text{th}$ question. NRM does not explicitly model the order of the categories. Instead, it learns two parameters $\beta_{jk} $ and $\alpha_{jk}$ for each category. The main limitation of NRM is interpretability \cite{thissen1997response}. In GPCM, the $\alpha$ values are ordered threshold values for choosing the next (more correct) category, and practitioners can use the $\alpha$ values to directly understand the ordering of the categories. In NRM, the concept of thresholds does not exist. NRM learns two interacting parameters $\alpha$ and $\beta$ (for each category) that jointly determine the ordering of categories. Unlike the ordered $\alpha$ values in GPCM, relative values of $\alpha$ in NRM do not provide intuitive ordering of the categories. 
 
\subsection{Existing Bayesian Ordinal Models}
\label{sec:ORD}

Assuming a known and fixed ordering of categories, the standard ordinal model \cite{odm99} posits a latent random variable $Z_{ij}^\prime, \forall (i,j) \in \Omega_{\text{obs}}$, defined as
\begin{align} \label{eq:ordmodelfun}
Z_{ij}^\prime = Z_{ij} + \varepsilon = \theta_i - \alpha_j + \varepsilon,
\end{align}
where $\varepsilon$ is assumed to be a standard normal random variable. The model further requires a set of ordered bin positions on the $j^\text{th}$ question denoted by $-\infty = \gamma^0_j < \gamma^1_j < \cdots < \gamma^{M_j} = \infty$, which map the latent predictor variable $Z_{ij}^\prime$ into one of the $M_j$ polytomous response categories as follows
\begin{align*}
Y_{ij} = y \quad \text{if} \quad \gamma_{j}^{y-1} <  Z_{ij}^\prime \leq \gamma_{j}^y. ~~~\forall{y \in \{1,\ldots,M_j\}}.
\end{align*}  
A common constraint imposed on the bin positions is $\gamma_j^1 = 0$ which avoids identifiability problems where the bin positions could be shifted and scaled arbitrarily \cite{odm99}. 

\begin{figure*}[t]
\begin{center}
\hfill
   \subfigure[ORD category density function]{\label{sprite_func}\includegraphics[width=0.25\columnwidth]{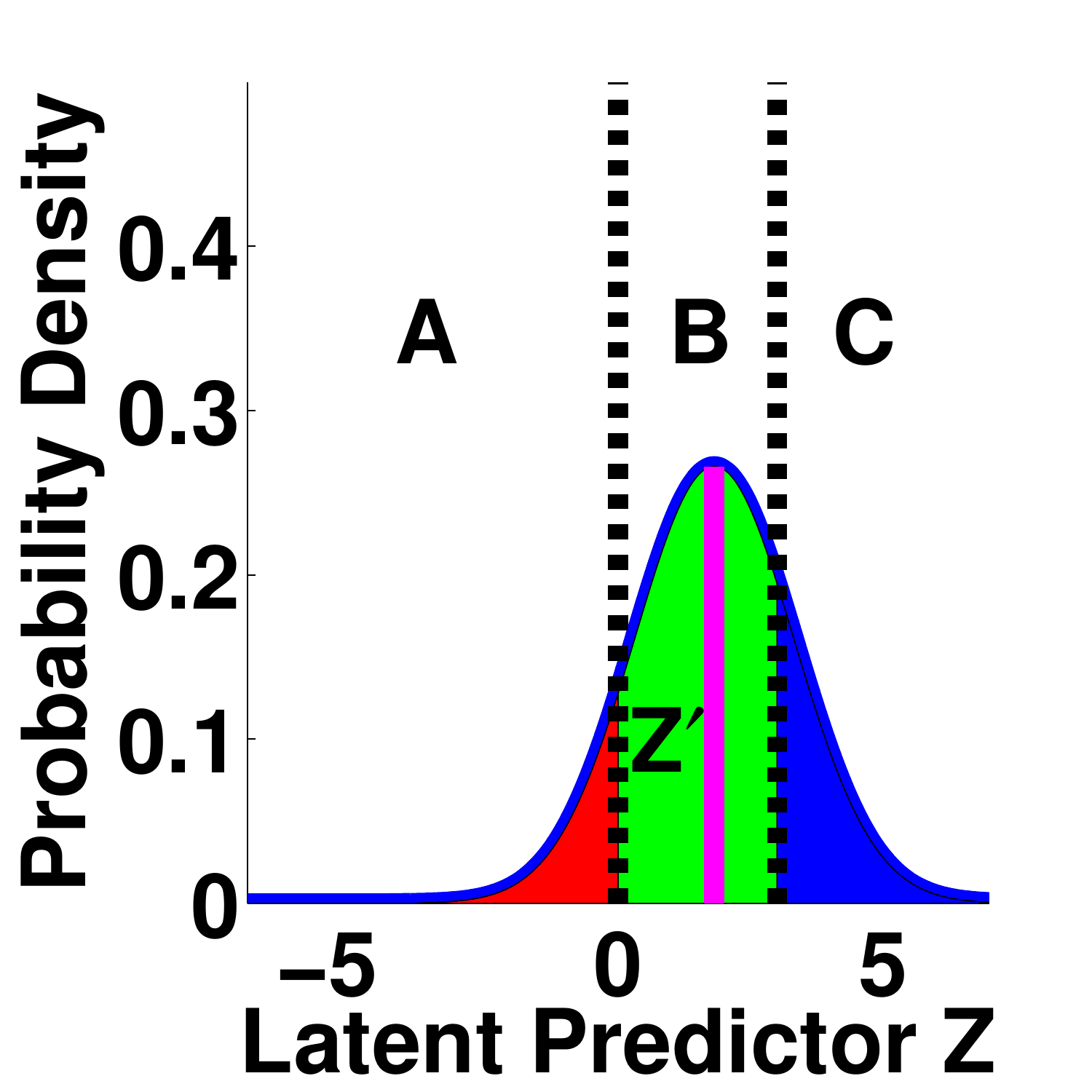}}\hfill
    \subfigure[ORD model]{\label{sprite_icrf}\includegraphics[width=0.25\columnwidth]{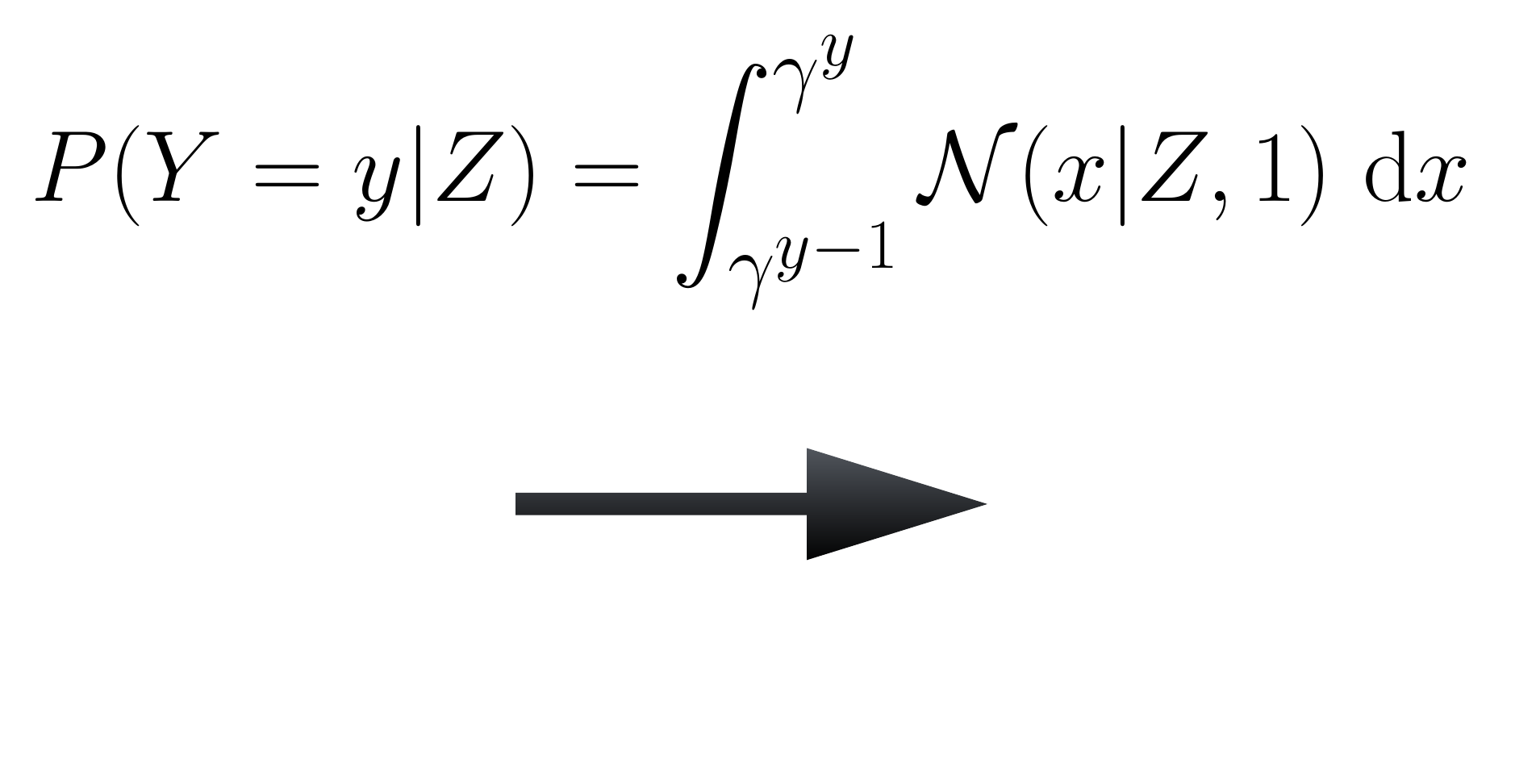}} \hfill
    \subfigure[Category probabilities at $Z^\prime$ ]{\label{sprite_bar}\includegraphics[width=0.25\columnwidth]{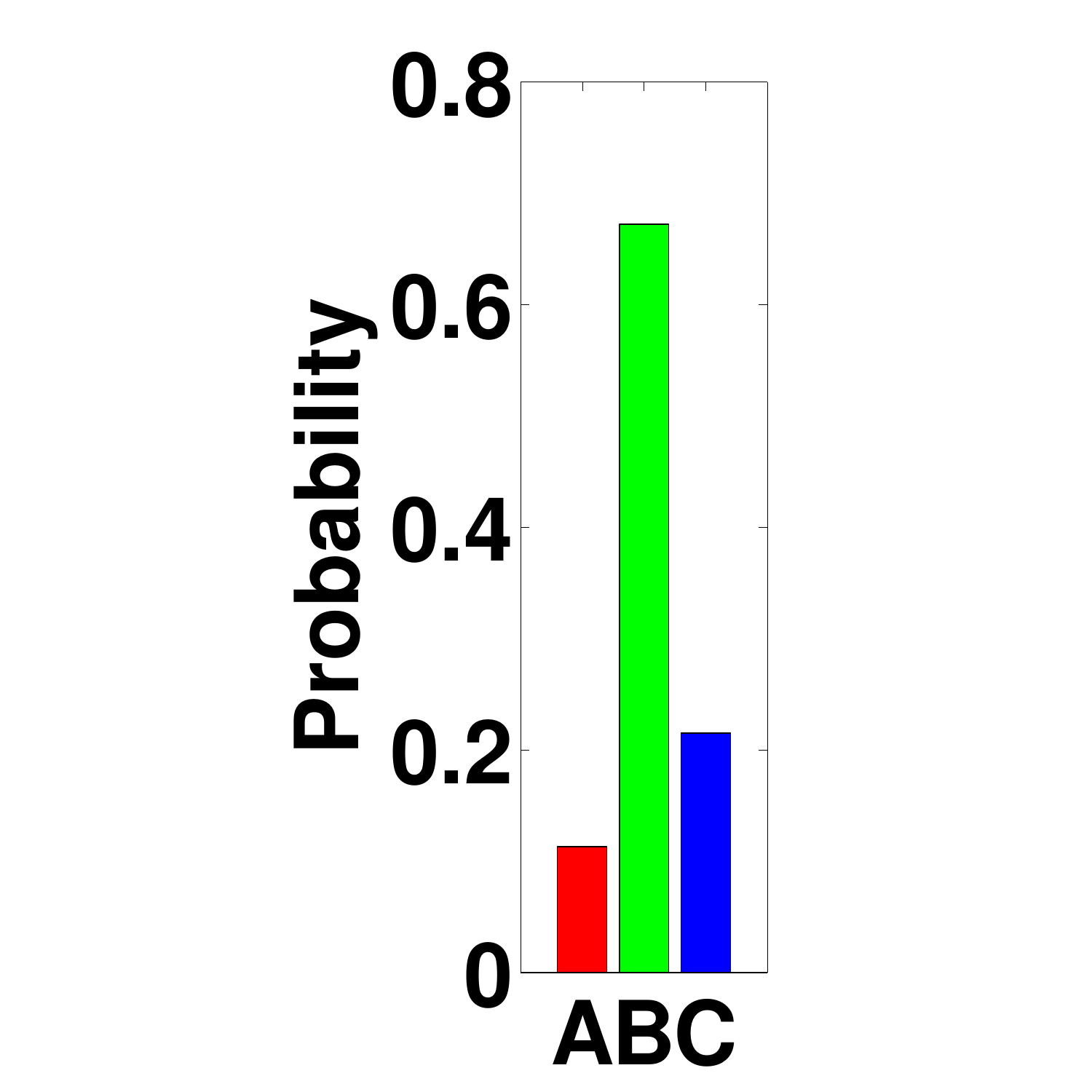}}\hfill
\end{center}
\caption{Illustration of the ORD model: The location of latent predictor variable $Z$ and the bin positions $\gamma$ induce a probability mass function determining the probability of category $y$ (out of A, B, and C) a respondent will choose. (a) $Z^\prime$ determines the mean of the Gaussian function shown, the dashed lines determine the category bin locations. (b) The ORD likelihood model. (c) Category probabilities at location $Z^\prime$.}
 \label{fig:ord_overview}
\end{figure*}

With \eqref{eq:ordmodelfun}, we can express the likelihood of selecting category $y \in \{1,\ldots,M_j\}$ as follows
\begin{align}
P(Y_{ij} = y | Z_{ij}, \boldsymbol{\gamma}_j) 
&= \Phi(\gamma^{y}_j \!-\! Z_{ij}) \!-\! \Phi(\gamma^{y-1}_j\! \!- Z_{ij}). \label{eq:likelihood_ordered}
\end{align}
Here, $\Phi(x)$ is the cumulative distribution function of the standard normal distribution,  defined as $\Phi(x) = \int_{-\infty}^x \frac{1}{\sqrt{2 \pi}} \text{exp}(\frac{-x^2}{2}) \text{d}x$. Figure \ref{fig:ord_overview} illustrates the ORD model.
As noted in the introduction, in a large number of practical applications the exact ordering of the categories is unknown a priori. Instead, we are faced with having only a set of $M_j$ $\textit{labels}$ for question $j$ and must $\textit{learn}$ the natural ordering of these labels from data. To be able to analyze unordered categorical data, we slightly modify ORD and introduce the learned ordinal (LORD) model that fuses the ORD model with a model on the ordering of the category labels. This modification requires one to learn a permutation $\pi$ that maps the $M_j$ labels into a new set of $M_j$ ordered values. There are $M_j!$ (where $M_j$ is the number of categories) such permutations which we denote by $\pi^\ell_j, \ell = 1,\ldots,M_j!$.

Given a specific permutation $\pi^\ell_j$, we can rewrite the generative likelihood of \eqref{eq:likelihood_ordered} as
\begin{align*}
P(Y_{ij} = y | Z_{ij}, \boldsymbol{\gamma}_j, \pi^\ell_j) = \Phi(\gamma^{\pi^\ell_j(y)}_j - Z_{ij}) - \Phi(\gamma^{\pi^\ell_j(y-1)}_j - Z_{ij}). 
\end{align*}
The prior distributions on each of the latent parameters of interest are given by
\begin{align}
\theta_i &\sim \mathcal{N}(0,\nu_\theta)			&  		\gamma_j &\sim \mathcal{N}(0,\nu_\gamma)  \notag 
\\
\alpha_j &\sim \mathcal{N}(0,\nu_\alpha)			&	  	\pi_j  &\sim \mathcal{U}[1,\ldots,M_j!], \notag
\end{align}
 where $\nu_\theta$, $\nu_\alpha$ and $\nu_\gamma$ are hyperparameters that define the prior variance for the latent respondent parameters, latent question parameters and the bin positions respectively. $\mathcal{N}$ and $\mathcal{U}$ represent the Gaussian distribution and uniform distribution respectively.%

\section{The SPRITE Model} 
\label{sec:sprite_model}
We now introduce our proposed SPRITE model. The models discussed in Section \ref{sec:ORD} combine the latent predictor $Z_{ij}$ with the bin positions $\boldsymbol{\gamma}_j$ in order to generate the observed response $Y_{ij}$. The SPRITE model, by contrast, does not rely on a set of bins, but rather on distributions over the categories themselves (see Figure \ref{fig:sprite_overview} for an illustration of this principle). For each question $j$, each category $k \in\{1,\ldots, M_j \}$ specifies a Gaussian function with mean $\mu_j^k$ and  variance $\nu_j^k$. We call each category's Gaussian function a "sprite". We model the probability that respondent $i$ will select category $y$ of question $j$ given the value of the latent predictor variable $Z_{ij}$ as follows
\begin{align*}
P(Y_{ij} = y | Z_{ij}) = \frac{\mathcal{N}(Z_{ij} | \mu_j^y, \nu_j^y)}{\sum_{k=1}^{M_j} \mathcal{N}(Z_{ij} | \mu_j^k, \nu_j^k)}.
\end{align*}
As described in Section \ref{sec:model_parameters}, $Z_{ij} = \theta_i - \alpha_j$. Since, given $\mu_j^y$, the parameter $\alpha_j$ does not offer any additional information, we will absorb the $\alpha_j$ terms directly into the mean $\mu_j^y$ and use $Z_{ij} = Z_{i} = \theta_i$ as the latent predictor. We therefore re-express the SPRITE likelihood equation using $Z_{i}$ as
\begin{align}
P(Y_{ij} = y | Z_{i}) = \frac{\mathcal{N}(Z_{i} | \mu_j^y, \nu_j^y)}{\sum_{k=1}^{M_j} \mathcal{N}(Z_{i} | \mu_j^k, \nu_j^k)}. \label{eq:NORD_likelihood}
\end{align}
 Figure \ref{fig:sprite_overview} (a) shows the item category density functions or "sprites" of each of the three categories induced over $Z$. Figure  \ref{fig:sprite_overview} (c) shows the item category response functions (ICRFs) that plots the probability of choosing each category as a function of $Z$.
 
The prior distributions on each of the latent parameter of interest are given by
\begin{align}
\mu_j^y & \sim \mathcal{N}(0,\nu_\mu)&
 Z_{i} & \sim \mathcal{N}(\mu_z,\nu_z) \notag \\
\nu_j^y & \sim \mathcal{IG}(\alpha_\nu,\beta_\nu) &
 y \notin \Omega_\text{obs}  & \sim \mathcal{U}[1,\ldots,M_j],\label{eq:NORD_prior}
\end{align}
where $\mathcal{IG(\alpha,\beta)}$ denotes the inverse gamma distribution with shape parameter $\alpha$ and scale parameter $\beta$, and $\nu_\mu$, $\alpha_\nu$, $\beta_\nu$, $\mu_z$, $\nu_z$ are hyperparameters for the prior distributions of the latent category mean, category variance, and respondent latent trait. We treat missing observations $y \notin \Omega_\text{obs}$ as latent parameters and use a uniform prior distribution on them. 
The associated inference details are given in Section \ref{sec:NORD_method}. \par Like all models for IRT used to analyze unordered categorical data, SPRITE can be susceptible to identifiability issues in the data. For example, one can negate all the learned categories means $\mu_j$ and at the same time negate the inferred respondent latent ability parameters $\theta_j$ without affecting the model likelihood. To prevent identifiability issues, we fix the mean of one sprite (typically the sprite whose category corresponds to the correct answer) to zero and its variance to one. SPRITE's modeling flexibility allows overlapping categories with no strict ordering. Furthermore, SPRITE's category mean and variance parameters offer superior interpretability compared to existing models.    

\newcommand{\facw}{0.35}

\section{Inference With SPRITE}
\label{sec:method}

We now detail our inference method for SPRITE.  We first note that, under the Bayesian setting, there exist a number of methods for fitting SPRITE to data.  We will rely on Markov Chain Monte-Carlo (MCMC) sampling methods \cite{gelman_book}, which are easy to deploy for our model. Unlike methods such as expectation maximization (EM) \cite{dempster1977maximum,bishop2006pattern} that produce point estimates, an MCMC-based approach provides full posterior distributions.  

\subsection{MCMC Sampler for SPRITE}
\label{sec:NORD_method}

We present a Metropolis-within-Gibbs sampler \cite{gilks1995adaptive} for SPRITE. The SPRITE latent variables $\boldsymbol{Z}$, $\boldsymbol{\mu}_j$, and $\boldsymbol{\nu}_j$ for $i =1,\ldots,N$ and $j=1,\ldots,Q$ are sampled via a Metropolis--Hastings step at each MCMC iteration. Here, we introduce the vector notation $\boldsymbol{\mu}_j = [\mu^1; \ldots; \mu^{M_j}]$, $\boldsymbol{\nu}_j = [\nu^1; \ldots ;\nu^{M_j}]$ and $\boldsymbol{Z} = [Z_1; \ldots; Z_N]$. Furthermore, we treat missing observations as latent variables and sample them using Gibbs sampling. A summary of the steps used by our MCMC sampler are as follows. We use the notation $[\cdot]^t$ to represent the state of a parameter at iteration number $t$. For $t = 1, \ldots, T$ where $T$ denotes the total number of MCMC iterations, we perform the following steps: 

\begin{enumerate}
\item  Propose new latent traits $[Z_i]^t \sim  \mathcal{N}\left([Z_i]^{t-1}, \nu_{z}\right)$ for $i = 1, \ldots, N$. 
\item Propose new category means ${[\mu_j^k]}^t \sim  \mathcal{N}\left({[\mu_j^k]}^{t-1}, \nu_{\mu_{}}\right)$ for $j = 1, \ldots, Q$ and $k=1,\ldots,M_j$.


\item Propose new category variances $[\nu_j^k]^t \sim \mathcal{IG}(\alpha_\nu,\beta')$, where the updated scale parameter $ \beta' = [\nu_j^k]^{t-1}(\alpha_v-1)$ for $j = 1, \ldots, Q$ and $ k=1,\ldots,M_j$. Note that the mean of $\mathcal{IG}(\alpha_v,\beta')$ is $[\nu_j^k]^{t-1}$.   

\item Calculate a Metropolis-Hastings acceptance/rejection probability based on the likelihood ratio between proposed parameters and the parameters from the previous MCMC step. The likelihood is given by \eqref{eq:NORD_likelihood} using $[Y_{ij}]^{t-1}$, $(i,j)\notin \Omega_\text{obs}$ and $Y_{ij}$, $(i,j)\in \Omega_\text{obs}$. The proposed latent variables  $[Z_i]^t$, ${[\mu_j^k]}^t$, and $[\nu_j^k]^t$ for $i = 1, \ldots, N$, $j = 1, \ldots, Q$ and $k=1,\ldots,M_j$ are then jointly accepted or rejected.

\item Propose new prediction values for the missing responses $[Y_{ij}]^t$, $(i,j) \notin \Omega_\text{obs}$ by Gibbs sampling the probabilities induced by \eqref{eq:NORD_likelihood} using $[Z_i]^t$, ${[\mu_j^k]}^t$,$[\nu_j^k]^t$.

\end{enumerate}
\noindent Above, $\nu_{z}$, $\nu_{\mu_{}}$, and $\alpha_{\nu_{}}$  are user-defined tuning parameters.

%
%
%

\subsection{Posterior Inference}
\label{sec:posterior_inference}
After a suitable burn-in period, the MCMC sampler detailed in Section~\ref{sec:NORD_method} produces samples that approximate the true posterior distribution of all model parameters.  We will make use of the posterior mean when performing experiments in which we compare to a known ground truth. For real data experiments with unknown ground-truth latent parameters, we gauge the performance of our models by measuring predictive accuracy (error metrics are presented in Section \ref{sec:real_exp_setup}). We make predictions using fully Bayesian imputation \cite{kong1994sequential} where we predict missing responses using the posterior mode of $Y_{ij}, (i,j) \notin \Omega_\text{obs}$.

%


\section{Experiments}
\label{sec:experiments}

We first evaluate SPRITE using synthetic data to demonstrate model convergence, identifiability, and consistency. Then, we compare the predictive performance of SPRITE to other IRT models (detailed in Section \ref{sec:model}) using real-world educational datasets.  

\subsection{Synthetic Data Experiments}

\paragraph{Generation of Data}
We first generate the ground truth SPRITE model parameters $\boldsymbol{Z}$, $\boldsymbol{\mu}_j$, and $\boldsymbol{\nu}_j$ for $i =1,\ldots,N$ and $j=1,\ldots,Q$. For simplicity, we fix the number of categories per question to $M_j = M = 5$, $\forall j$. We generate the latent parameters via \eqref{eq:NORD_prior} and the observed data $\mathbf{Y}$ via \eqref{eq:NORD_likelihood}. In this experiment, the graded response matrix $\boldsymbol{Y}$ is assumed to be fully observed. The hyperparameters are as follows: $\mu_{z}=0$, $\nu_{z} =1$, $\nu_{\mu}=1$, $\alpha_{\nu}=1,$ and~$\beta_{\nu}=1$. 

\paragraph{Parameter estimation and error metrics}

We deploy SPRITE as described in Section \ref{sec:NORD_method} by  initializing all parameters of interest with random values.  We use 90,000 MCMC iterations in the burn-in phase and compute the posterior means for all parameters as described in Section \ref{sec:posterior_inference} over an additional 10,000 iterations.
We compare the learned \NORD~parameters to the known ground truth model using the following three error metrics
\begin{align}
\quad E_{\boldsymbol{Z}} \!=\! \frac{\|\hat{\boldsymbol{Z}}-{\boldsymbol{Z}}\|^2}{\|{\boldsymbol{Z}}\|^2}, \,
\quad E_{\boldsymbol{\mu}} \!=\! \frac{\|\boldsymbol{\hat\mu}-\boldsymbol{\mu}\|^2}{\|\boldsymbol{\mu}\|^2}, \, 
\quad E_{\boldsymbol{\nu}} \!=\! \frac{\|\boldsymbol{\hat\nu}-\boldsymbol{\nu}\|^2}{\|\boldsymbol{\nu}\|^2},\label{eq:errormets}
\end{align}
where $\hat{\boldsymbol{Z}}$, $\hat{\boldsymbol{\mu}}$, and $\hat{\boldsymbol{\nu}}$ represent model estimated values as computed in Section \ref{sec:NORD_method} and $\boldsymbol{Z}$, $\boldsymbol{\mu}$, and $\boldsymbol{\nu}$ represent the known ground-truth values. 
\paragraph{Discussion}

\begin{figure*}[ht]
  \begin{center}
    \subfigure[Respondent latent trait error $E_{\boldsymbol{Z}}$.]{\label{synth_c}\includegraphics[width=0.35\columnwidth]{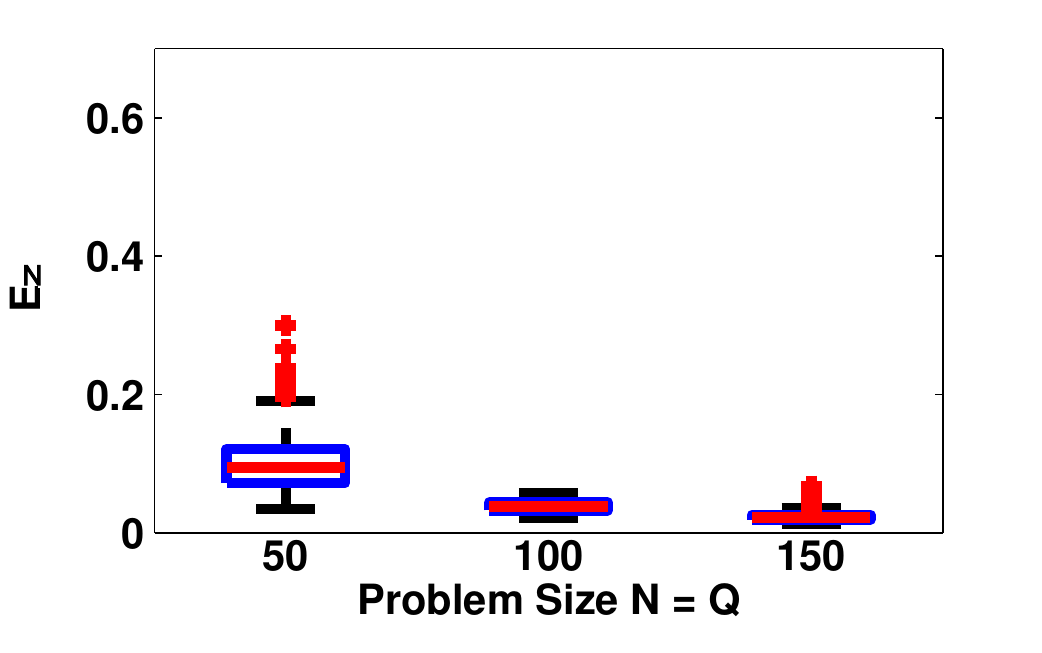}}
    \subfigure[Category mean error $E_{\boldsymbol{\mu}}$.]{\label{synth_optm}\includegraphics[width=0.35\columnwidth]{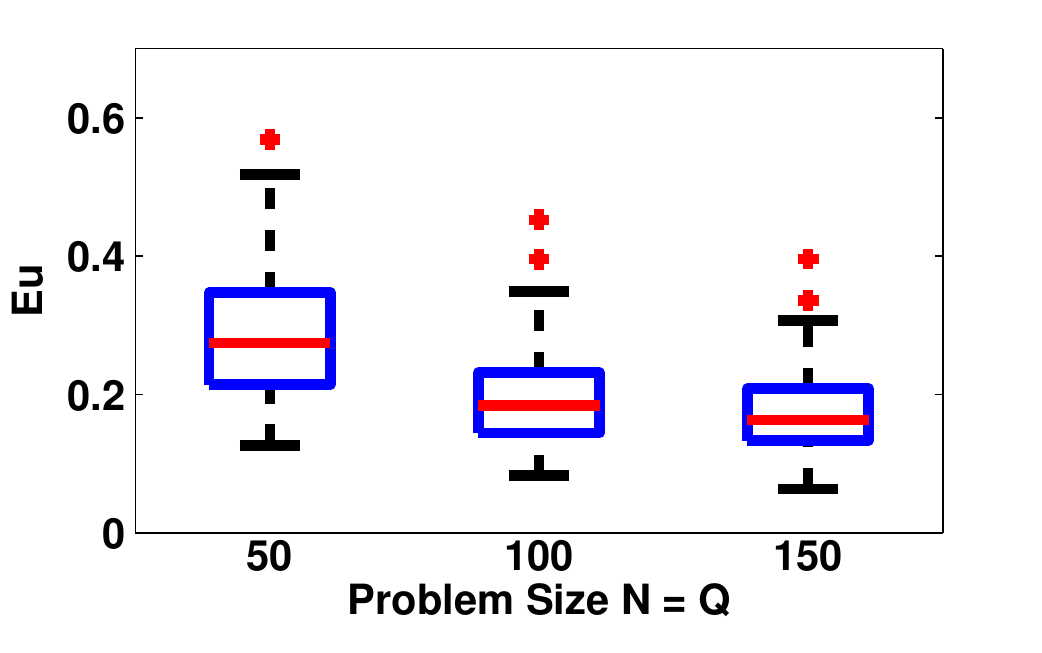}} 
    \subfigure[Category variance error $E_{\boldsymbol{\sigma}}$.]{\label{synth_optv}\includegraphics[width=0.35\columnwidth]{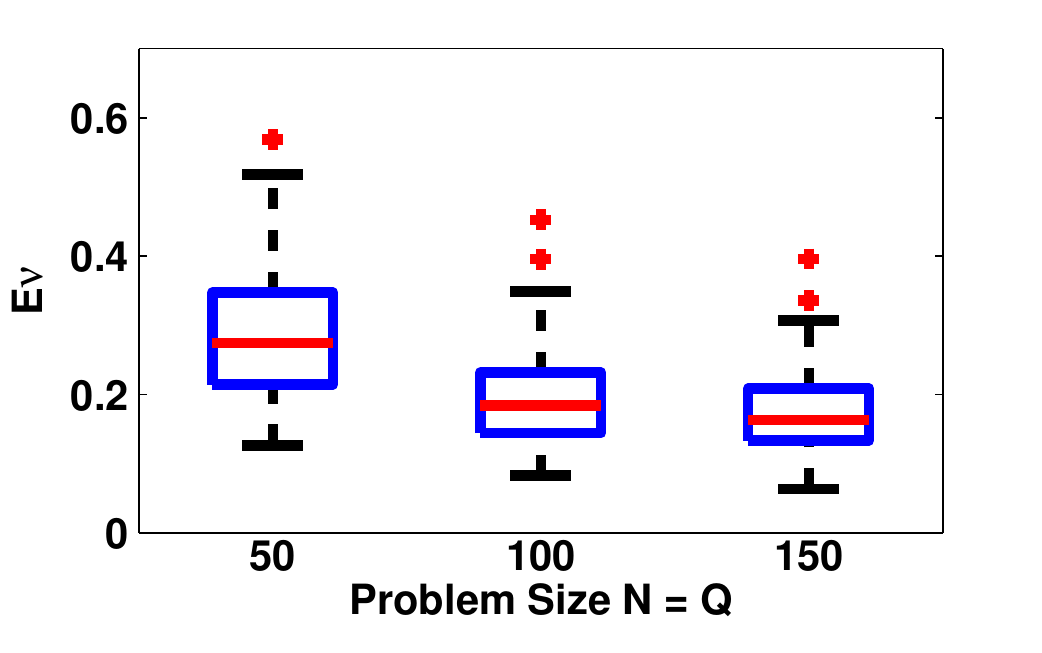}}
  \end{center}
  \caption{Synthetic experiment over various problem sizes (number of respondents $N$ and number of questions $Q$) where $N$ = $Q$, and $M_j = M = 5$ categories. (a) The error of the latent trait $E_{\boldsymbol{Z}}$; (b) The error in the means of the categories $E_{\boldsymbol{\mu}}$; (c) The error of category variance $E_{\boldsymbol{\sigma}}$. All three error metrics decrease as the problem size (the amount of data) grows.}
  \label{fig:synth}
\end{figure*}

Figure \ref{fig:synth} displays box-whisker plots for the 3 error metrics in \eqref{eq:errormets} for various problem sizes (number of questions and number of respondents). To simplify the presentation of results, we keep the number of questions and number of respondents the same for all problem sizes. The low error rates demonstrate SPRITE model identifiability and its convergence to the ground truth. The low standard deviation values demonstrate SPRITE model stability.  Furthermore, all error metrics decrease as the problem size increases, which implies model consistency.

\subsection{Real-World Data Experiments}
\label{sec:ed_exp}

We now compare the predictive performance of SPRITE against the NRM, the GPCM, and the ORD methods (described in Section \ref{sec:model}) using a variety of real-world datasets. We use the LORD method outlined in Section~\ref{sec:ORD} in place of ORD when the category ordering is unknown a priori. 
\paragraph{Dataset details}

We study five educational datasets. A brief description of the datasets can be found in Table \ref{tbl_datasets}. The ``algebra test'' dataset is from a secondary level algebra test administered on Amazon's Mechanical Turk \cite{lan2013sparse}.  All questions are multiple choice questions and a domain expert has provided an ordering to the multiple choice categories (according to the correctness of each category). The datasets ``computer engineering course,'' ``probability course,'' and ``signals and systems course'' are from college level courses administered on OpenStax Tutor \cite{stax}. Each of these datasets contain a number of missing entires---corresponding to the case where students did not answer all available questions.
Finally, the ``comprehensive university exam'' dataset contains responses on an university level comprehensive exam \cite{vats2013test}. There are missing entries in this dataset because students were penalized less for choosing to skip a question instead of answering incorrectly. There is no a priori category ordering knowledge for all datasets except for the ``algebra test'' dataset, where a human expert has provided category ordering. 

\begin{table}
  \caption{Description of datasets. Unobserved data listed in the table refers to actual missing responses in the respective datasets; $Q$ denotes the number of questions in each dataset and $N$ denotes the number of respondents.}\vspace*{1ex}
  \label{tbl_datasets}
  \centering
  \begin{tabular}{llccc}
    \toprule
     \bf Description & {\bf Size} ($Q\times N$)& \bf Categories & \bf Ordered & \bf Observed data\\
    \midrule
Algebra test &34 $\times$ 99 &5 &Yes  &$100$\%\\
Computer engineering course & 203 $\times$ 82 & 12 &No  &$97$\%\\
Probability course & 86 $\times$ 49 & 7 &No &$67$\%\\
Signals and systems course & 143 $\times$ 44 & 11  &No  &$64$\%\\
Comprehensive university exam & 60 $\times$ 1567 & 4 &No  &$71$\%\\
    \bottomrule
  \end{tabular}
\end{table}

\paragraph{Experiment setup} 
\label{sec:real_exp_setup}

We compare the predictive performance of the algorithms by first puncturing (removing) a portion of the observed data and retaining these values for a test set. We set the rate of puncturing to be 20\%. We then train each model using the remaining observed entries and make predictions on the test set. Once the models have been fit, we infer the missing entries as discussed in Section \ref{sec:posterior_inference}. The error metric used in all educational datasets is simply the number of incorrect predictions divided by the total number of predictions made. All experiments were repeated over $50$ random puncturing patterns. We use 90,000 MCMC sampling iterations for the burn-in period and compute our results over an additional 10,000 iterations.

\paragraph{Results and discussion} 
The predictive performance results of all models are summarized in Table \ref{tbl_results}. SPRITE outperforms all other models on all datasets. The "algebra test" dataset is especially interesting, where a human expert has provided a strict ordering of the categories. This expert provided ordering was used by ORD, which requires a priori known category ordering. SPRITE, on the other hand, learned a category ordering directly from the data without considering the one provided by the human expert. Compared to the category ordering provided by the human expert, SPRITE's learned ordering is more flexible (allowing overlapping categories). Furthermore, SPRITE's learned ordering is completely data driven and is not influenced by the human expert's subjective opinion, which is often unreliable. SPRITE's superior performance in the "algebra test" dataset demonstrates that the SPRITE learned category ordering explains the data better than the one provided by the human expert. 
\par
These experiments show that SPRITE performs well against other IRT models on both ordered and unordered categorical data. Furthermore, SPRITE often learns superior category orderings than the ones provided by human experts.

\paragraph{Interpretability and mutual information}

\sloppy

SPRITE's category parameters $\boldsymbol{\mu}_j$ and $\boldsymbol{\nu}_j$ provide an intuitive ordering of categories. Furthermore, SPRITE provides valuable statistics concerning question informativeness. In the context of education, the categories chosen by each learner provide information about their particular mastery of the material.  Similarly, the learners inform \NORD~about how well each question/category discriminates learners with strong vs. \ weak mastery of the material. 

In particular, using the statistics provided by SPRITE, we can compute the mutual information $I(Z;Y_{j})$ (measured in bits) between the learners' latent abilities $Z \in \mathbb{R} $ and the category choices~$Y_{j} \in 1,\ldots,M_j$ made for each question $j = 1,\ldots,Q$. 
%
%
The mutual information (MI) is able to reveal the informativeness or discriminative power for a given question $j$. The MI is defined as follows %
\fussy
%
%
\begin{align}
 I(Z;Y_j) &= \sum^{M_j}_{y=1} \int_{\mathbb{R}}{\text{P}(y|z)\text{P}(z)\log_2{\frac{\text{P}(y|z)}{\text{P}(y)}} \text{d}z}. 
\label{eq:mi}
\end{align}
Here, $\text{P}(y|z)$ is the likelihood function given by \eqref{eq:NORD_likelihood}, $\text{P}(z) = \mathcal{N}(z | \mu_z,\nu_z)$ is the Gaussian prior on latent abilities given by \eqref{eq:NORD_prior}, and P$(y)=\int_{\mathbb{R}}{\text{P}(y|z)\text{P}(z)\text{d}z} $ is a normalization term. 
The integral in  \eqref{eq:mi} is difficult to evaluate in closed-form. However, it can be evaluated easily and accurately using numerical integration techniques. 

\begin{figure}
\centering

\hspace*{\fill}%
   \subfigure[An informative question with MI of $0.42$\,bit]{\label{himi}\includegraphics[width=0.35\columnwidth]{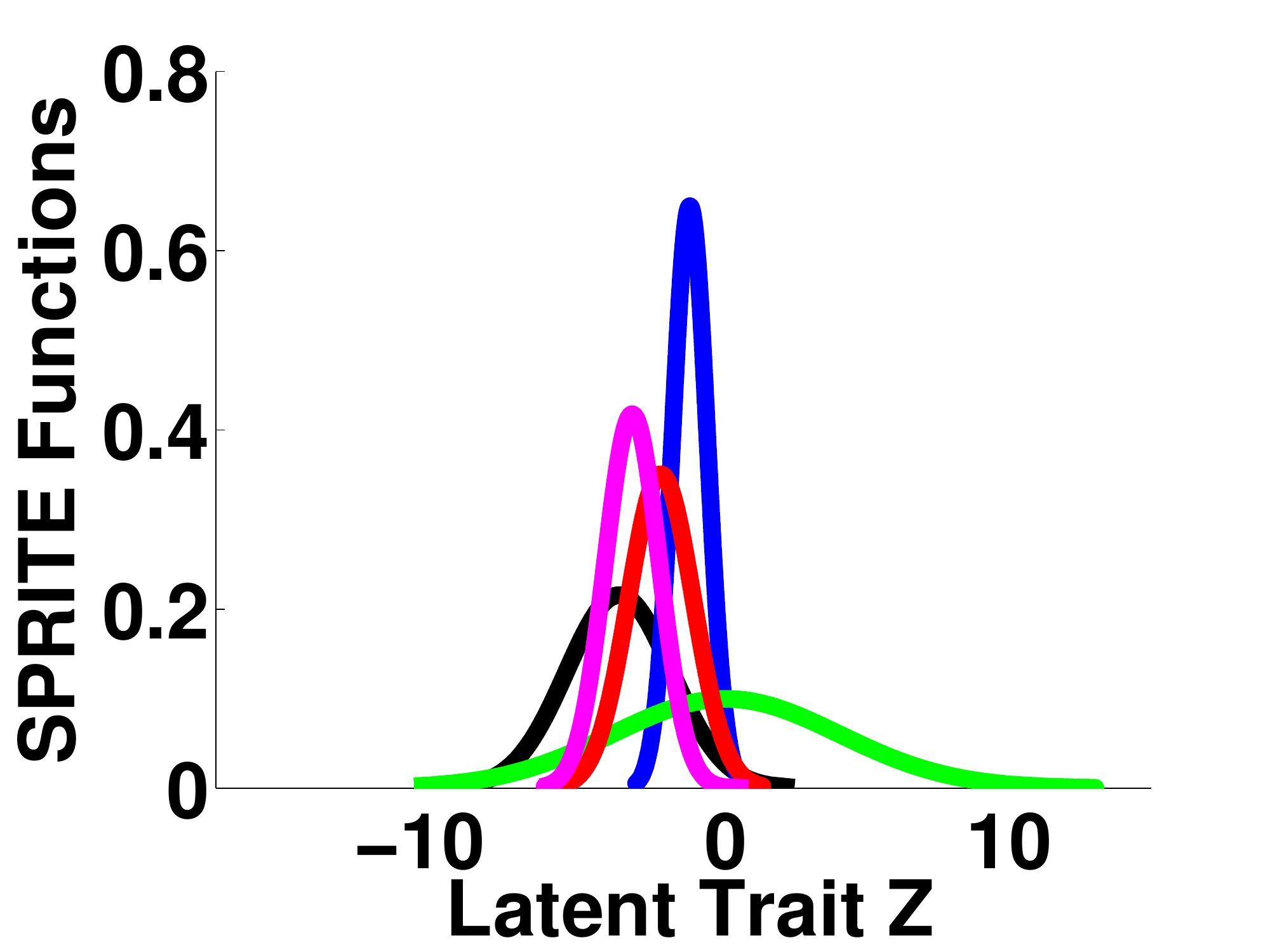}}\hspace*{\fill}%
    \subfigure[A un-informative question with MI of $0.08$\,bit]{\label{lowmi}\includegraphics[width=0.35\columnwidth]{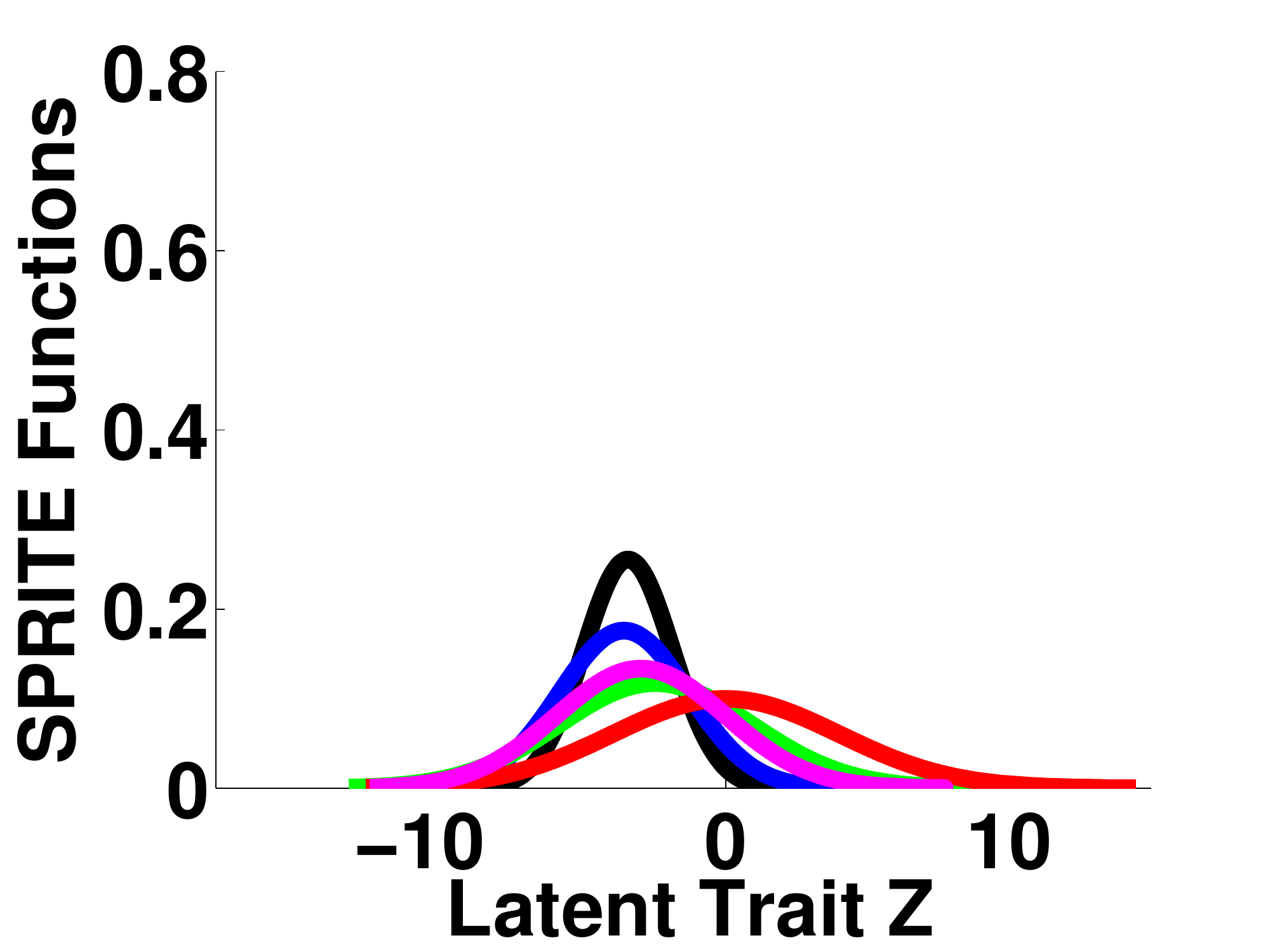}}
\hspace*{\fill}%

%
%
\caption{ Learned parameters of two questions using SPRITE from the ``algebra test'' dataset. The curves represent SPRITE category functions or sprites. The colors of the curves have no meaning and are only used to aid visual diambiguation of unique sprites. (a) Shows an informative question with $\text{MI} = 0.42$\,bits. (b) Shows a less informative question with $\text{MI} = 0.08$\,bits.}
 \label{fig:turk_analytics}
\end{figure}

 %
%
%
%
%
%
Figure \ref{fig:turk_analytics} demonstrates the efficacy of the MI measure \eqref{eq:mi} for one informative and one less informative question in the ``algebra test'' dataset. The informative question ($\text{MI} = 0.42$\,bit) illustrated in Figure \ref{fig:turk_analytics} (a) reveals that one sprite dominates in the positive $Z$ region. This means that one category is able to distinguish the learner's performance very well from the other four categories. By contrast, the less informative question  ($\text{MI}= 0.08$\,bit) illustrated in Figure \ref{fig:turk_analytics} (b) reveals multiple overlapping sprites that show little discriminative power (all sprites are grouped fairly closely). In other words, the categories in this question fail to discriminate each learner's latent understanding. Instructors can use such information to either improve the quality of the available test questions (by revising certain categories) or to determine a high-quality subset of questions, which is key for test-size reduction \cite{vats2013test}.

%
%

\section{Conclusion}
\label{sec:conclusion}

We have developed SPRITE to model both ordered and unordered categorical response data. SPRITE improves upon the state-of-the-art IRT models both in interpretability and data fitting. Additionally, SPRITE provides valuable statistics regarding questions and categories (such as their efficacy and degree of information) that can be used to improve the quality of the test questions.
\sloppy
Several future directions look promising. First, improvements to the SPRITE sampler could potentially improve the efficiency of the SPRITE inference algorithm.  Methods such as variational Bayes \cite{attias1999inferring}, expectation maximization \cite{bishop2006pattern} and Metropolis-Hastings Robbins-Monro \cite{cai2010metropolis} may sacrifice little in terms of data fitting performance while providing improvements in computational time. Additionally, alternative models for the linear predictor $Z$, such as MIRT \cite{beguin2001mcmc} and linear regression models with either fixed or learned covariates, could easily provide additional improvements in terms of performance and interpretability. 
\fussy   

\section*{Acknowledgments}

Thanks to Mr.~Lan for discussions on the SPRITE model and Charles Jeon for his help with numerically computing the mutual information. The authors would like to express their gratitude to the Chairman, JAC, IISER Pune, for sharing the "comprehensive university exam" dataset, as well as Divyanshu Vats for insightful discussion regarding this dataset.

\FloatBarrier

\bibliographystyle{acmtrans}
\bibliography{NORD}

\end{document}